\pdfoutput=1
\documentclass[11pt]{article}

\usepackage[final]{acl}

\usepackage{times}
\usepackage{latexsym}

\usepackage[T1]{fontenc}
\usepackage[utf8]{inputenc}

\usepackage{microtype}

\usepackage{inconsolata}

\usepackage{graphicx}
\usepackage{amsmath}
\usepackage{booktabs}
\usepackage{multirow}
\usepackage{makecell}
\usepackage{tcolorbox}
\usepackage{algorithm}
\usepackage{algpseudocode}
\title{From Long Videos to Engaging Clips: A Human-Inspired Video Editing Framework with Multimodal Narrative Understanding}

\newcommand*{\affaddr}[1]{#1}
\newcommand*{\affmark}[1][*]{\textsuperscript{#1}}
\newcommand*{\email}[1]{\texttt{#1}}

\author{
Xiangfeng Wang\thanks{Equal contribution.}\affmark[\textnormal{1}], 
Xiao Li\footnotemark[1]\affmark[\textnormal{2}], 
Yadong Wei\affmark[\textnormal{2}], 
Xueyu Song\affmark[\textnormal{2}], 
Yang Song\affmark[\textnormal{2}], \\
\textbf{Xiaoqiang Xia}\affmark[\textnormal{2}], 
\textbf{Fangrui Zeng}\affmark[\textnormal{2}], 
\textbf{Zaiyi Chen}\affmark[\textnormal{2}], 
\textbf{Liu Liu}\affmark[\textnormal{2}], 
\textbf{Gu Xu}\affmark[\textnormal{2}], 
\textbf{Tong Xu}\thanks{Corresponding author.}\affmark[\textnormal{1}] \\
\affaddr{\affmark[1]University of Science and Technology of China} \\
\affaddr{\affmark[2]ByteDance China} \\
\email{xf9462@mail.ustc.edu.cn, tongxu@ustc.edu.cn} \\
\email{\{lixiao.dlut, weiyadong, songxueyu, songyang.0708, xiaxiaoqiang\}@bytedance.com}\\
\email{\{zengfangrui.handsome,chenyi.cycy, liuliu.qt,xudongsheng.2020\}@bytedance.com} \\
\url{https://huggingface.co/datasets/wonderful9462/DramaAD} \\
}

\begin{document}
\maketitle

\begin{abstract}

The rapid growth of online video content, especially on short video platforms, has created a growing demand for efficient video editing techniques that can condense long-form videos into concise and engaging clips. 
Existing automatic editing methods predominantly rely on textual cues from ASR transcripts and end-to-end segment selection, often neglecting the rich visual context and leading to incoherent outputs. 
In this paper, we propose a \textbf{H}uman-\textbf{I}nspired automatic \textbf{V}ideo \textbf{E}diting framework (HIVE) that leverages multimodal narrative understanding to address these limitations. 
Our approach incorporates character extraction, dialogue analysis, and narrative summarization through multimodal large language models, enabling a holistic understanding of the video content. 
To further enhance coherence, we apply scene-level segmentation and decompose the editing process into three subtasks: highlight detection, opening/ending selection, and pruning of irrelevant content. 
To facilitate research in this area, we introduce DramaAD, a novel benchmark dataset comprising over 2500 short drama episodes and 500 professionally edited advertisement clips. 
Experimental results demonstrate that our framework consistently outperforms existing baselines across both general and advertisement-oriented editing tasks, significantly narrowing the quality gap between automatic and human-edited videos.
\end{abstract}

\section{Introduction}
In recent years, the rapid increase of online video content has significantly enriched users’ viewing experiences. 
However, it has also increased the cognitive burden of content consumption, particularly for long-form videos. 
Video editing—specifically, condensing lengthy videos with low information density into short, engaging, and information-rich clips—offers a potential solution. 
Such edited clips not only alleviate users’ browsing pressure but also facilitate easier content sharing. 
With the rise of short video platforms, the demand for efficient video editing has grown substantially. 
Currently, video editing is largely performed manually by content creators, who must watch the entire video and identify highlights for re-creation. 
Although manually edited clips tend to be of higher quality, this process is labor-intensive and inefficient, making it unsuitable for large-scale applications. 
As a result, automatic video editing has emerged as a promising solution to address these challenges.

% \begin{figure}
%   \centering
%   \includegraphics[width=1.0\linewidth]{image/intro.png}
%   \vspace{-1.em}
%   \caption{Distinctions between common automatic video editing methods and human editors. When editing videos, human editors primarily focus on the narrative and audience experience, while prior methods typically do not consider these aspects. Inspired by human editors, our approach similarly prioritizes the video's narrative and audience experience.}
%   \label{fig:intro}
%   \vspace{-1.em}
% \end{figure}

Early approaches to automatic video editing typically adopt end-to-end frameworks for training editing models. 
Some studies \cite{chen2016learning, podlesnyy, ave} formulate the task as a sequence selection problem, effectively performing subtraction on the original video by removing less informative segments. 
Other works \cite{hu2023reinforcement} treat it as a next-shot retrieval problem, performing addition by selecting and assembling shots to construct the edited video. 
However, these early methods were limited by the capacity of the models and lacked explicit understanding of narrative structures within videos. 
Consequently, the generated clips often failed to exhibit coherent storylines or logical progression.

In recent years, the emergence of large language models (LLMs) and multimodal large language models (MLLMs) has significantly advanced the capability for video narrative understanding, making content-aware automatic video editing increasingly feasible.
Existing LLM-based editing approaches, such as FunClip \footnote{\url{https://github.com/modelscope/FunClip}} and AI YouTube Shorts Generator \footnote{\url{https://github.com/SamurAIGPT/AI-Youtube-Shorts-Generator}}, typically employ automatic speech recognition (ASR) to transcribe spoken dialogue into text, which is then processed by LLMs to infer the narrative and select segments for editing.

However, these ASR-based methods primarily rely on dialogue and thus lack access to the rich visual information present in videos. 
Thus, they often fail to capture non-verbal cues such as facial expressions, gestures, and other visual context, leading to an incomplete understanding of the video content.
In addition, we observe that existing methods tend to adopt relatively simplistic editing strategies. 
They often rely on prompt engineering to elicit end-to-end predictions from the model regarding which segments should be included in the final clip. 
However, this approach frequently results in abrupt transitions between shots, which can negatively affect the overall viewing experience. 
In summary, automatic video editing faces two major challenges: (1) an incomplete understanding of video narratives caused by insufficient fusion of visual and textual modalities, and (2) the absence of systematic editing strategies, often leading to abrupt transitions and incoherent storytelling.

To address these challenges, we propose HIVE, an automatic video editing framework based on multimodal narrative understanding. 
The HIVE considers both the visual content and the underlying storyline, enabling the generation of diverse and engaging edited videos. 
Specifically, to achieve accurate video comprehension, we incorporate not only character extraction and dialogue analysis but also leverage MLLMs to generate narrative summaries of video content, thereby compensating for missing visual information. 
To enhance the coherence and viewing experience of the edited clips, we adopt an improved video scene segmentation \cite{scene_segmentation}, instead of conventional shot detection \cite{shot_detection, transnetv2}, to divide the video into semantically complete scenes. 
This ensures that clip boundaries do not interrupt ongoing dialogue or character actions.

% Furthermore, to explore a more effective editing approach, we consult professional editors for common manual editing techniques. 
% Inspired by their insights, we decompose the editing task into three relatively simple sub-tasks: highlight localization, opening/ending selection, and irrelevant content pruning. 
% This step-by-step approach allows LLMs to refine the video incrementally, mirroring human workflows.
% We observe that this method produces more natural and stable results compared to end-to-end generation.
Furthermore, to achieve human-level editing quality, we conducted a detailed study of the editing techniques commonly employed by professional editors. 
Inspired by their strategies, we decompose the editing task into three relatively simple sub-tasks: highlight detection, opening/ending selection, and irrelevant content pruning. 
This decomposition allows the LLM to progressively refine the video in a manner analogous to human editors. 
Our experimental results show that this step-by-step approach leads to more coherent and polished results compared to end-to-end editing methods.

% To advance research in automatic video editing, we introduce DramaAD, a benchmark dataset built from real short drama data. 
% Short dramas are fast-paced serialized videos, typically lasting 1\textasciitilde 3 minutes per episode, and have recently gained popularity on short video platforms.
% In practice, companies may seek to use automatic editing to transform short dramas into advertisement clips, thereby reducing labor costs.
% Our dataset offers both research and practical value for this emerging application scenario.
% DramaAD contains over 800 short drama videos and 500+ manually edited advertisement videos as references.
% To enable a relatively objective evaluation of edited videos, we introduce a set of quantitative metrics from multiple dimensions. 
% Experimental results show that our method significantly outperforms the baselines on both advertisement and general video editing scenarios, while narrowing the gap to human editors.

To support research in automatic video editing, we present DramaAD, a novel benchmark dataset constructed from real-world short drama videos. 
These short dramas, fast-paced serialized episodes typically lasting 1 to 3 minutes, have recently surged in popularity on short video platforms. 
In practical applications, companies are increasingly interested in repurposing such content into advertisement-style clips through automatic editing, aiming to reduce manual effort and production costs. 
DramaAD addresses this emerging need by offering both academic and industrial value: it comprises over 2500 raw drama videos and more than 500 professionally edited advertisement clips as references. 
To facilitate objective evaluation, we also propose a suite of quantitative metrics covering multiple aspects of editing quality. 
Experimental results demonstrate that our proposed framework significantly outperforms existing baselines in both ad-specific and general video editing tasks, while closing the performance gap with human editors.

Our main contributions are as follows: 
\begin{itemize}
    \item We introduce a multimodal narrative understanding approach that comprehensively interprets video content, ensuring smooth and engaging edited videos.
    
    \item We propose an innovative human-inspired automatic video editing framework that decomposes the editing task into three LLM-friendly sub-tasks, enabling superior editing performance.

    \item To support research in automatic video editing, we have created a dataset of short dramas in the advertising domain, which includes manually edited videos for reference. 
\end{itemize}

\section{Related Work}

\subsection{Automatic Video Editing}
In recent years, automatic video editing has gained increasing attention \cite{soe2021ai, ronfard2021film,koorathota2021editing,wang2019write}. 
Early methods primarily relied on feature extraction for video editing.
\citet{pardo2021learning} trained a ranking model using contrastive learning techniques, which can rank and identify plausible cuts among all possible transitions between a given pair of shots.
\citet{podlesnyy} designed a data-driven editing model that treats the video editing task as a sequence decision problem, allowing the model to determine whether each shot should be retained.
\citet{hu2023reinforcement} utilized a pretrained Vision-Language Model to extract editing-related representations and train an editor model by reinforcement learning.
These methods lack an explicit understanding of the video content and suffer from a lack of interpretability.
Recently LLM-based editing methods have made progress.
Some works \cite{barua2025lotus, lave} utilize LLMs to assist human editors, while other fully automated approaches, such as FunClip, leverage ASR-transcribed speech to achieve video editing.

Compared to these methods, our approach provides a more comprehensive handling of videos and superior editing strategies.

\subsection{Video Understanding}
Recently, significant progress has been made in LLM-based video understanding. 
Methods such as VideoChat2 \citep{videochat2} and LLaVA-Video \citep{llavavideo} integrate video foundation models with LLMs.
VideoChat2 uses a learnable Q-former inspired by BLIP2 \citep{blip2} for aligning visual and textual features, while LLaVA-Video employs a simple linear projection trained jointly on image-video data.  
Other methods, such as ChatVideo \citep{chatvideo} and VideoAgent \citep{videoagent}, incorporate LLMs within agent frameworks. 
% ChatVideo represents videos by trajectories and their attributes to support LLM-based reasoning. 
% VideoAgent relies on iterative analysis, dynamically selecting keyframes and using multimodal reasoning to understand videos.
Furthermore, general-purpose multimodal models, such as GPT-4o \citep{gpt-4o} and Gemini \citep{gemini}, also demonstrate powerful visual understanding abilities applicable to video tasks. 
In addition, Some studies focus on understanding films through audio description \citep{autoad}, narration \citep{mm}, and storytelling \citep{storyteller}, requiring finer video understanding and stronger narrative flow. 
In the short-video domain, datasets such as SkyScript-100M \citep{skyscript} provide structured textual formats capturing scenes, emotions, and cinematography instructions, thus supporting AI-based video generation tasks. 
Similarly, Vript \citep{vript} creates detailed mappings between general videos and fine-grained screenplay text, enabling more precise video description and understanding.

In contrast to these works, our research focuses specifically on video semantic analysis and leverages multi-dimensional information to achieve deeper video understanding.
% We integrate multi-dimensional information including character attributes, interpersonal relationships, and scene context to achieve deeper video understanding and content analysis.

\section{Method}
% \subsection{Task Definition}
% Given a video $V$, the automatic video editing based on narrative understanding first segments $V$ into sequential clips $(v_1,v_2,...,v_n)$ according to their semantics and generates coherent storyline descriptions $T=(t_1,t_2,...,t_n)$.
% Subsequently, the editing algorithm $f$ uses this information to produce the edited video $V'=f(V,T)=(v'_1,v'_2,...,v'_m)$, where $m\leq n$ and $v'\in V$.
\begin{figure*}[!t]
  \centering
  \includegraphics[width=\linewidth]{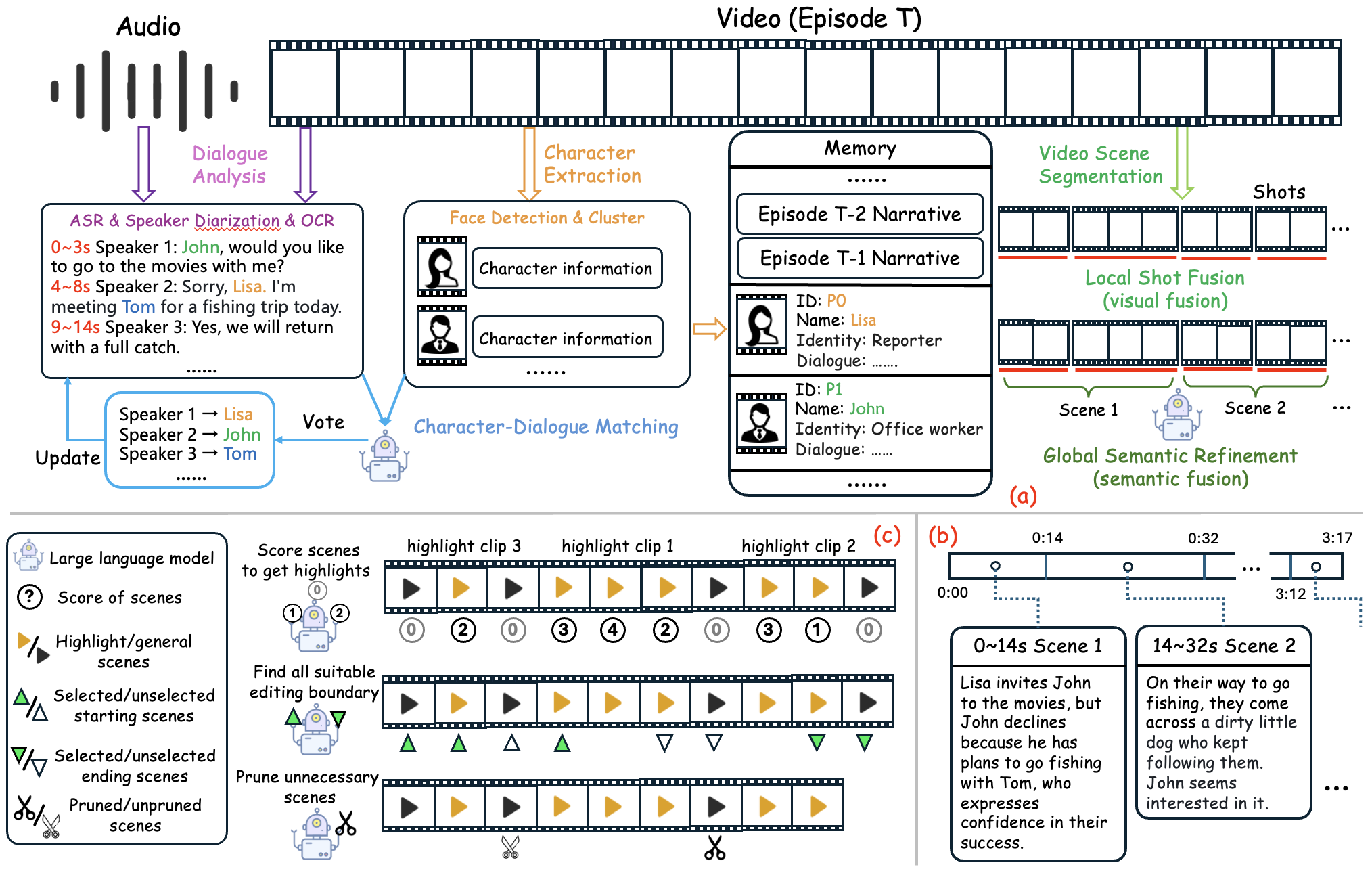}
  \vspace{0.em}
  \caption{An overview of the proposed framework. The video understanding module \textbf{(a)} extracts video information across multiple dimensions, such as dialogues and characters, etc., and segments the video into semantically complete scenes. Consequently, we obtain a sequence of scenes along with their corresponding narrative descriptions \textbf{(b)}. This content is then fed into the editing module \textbf{(c)}, which identifies highlights, selects appropriate editing boundaries, and removes unnecessary scenes as human editors do (take highlight clip 1 for example).}
  \label{fig:method}
  \vspace{-1.em}
\end{figure*}

To tackle automatic video editing, we propose a two-stage framework:
(1) Comprehensive content analysis, including characters, dialogues, character-dialogue matching and video scene segmentation, to obtain accurate narratives for each scene.
(2) An editing strategy inspired by human editors, which detects highlight clips, selects attention-grabbing openings \& suspenseful endings and prunes redundant content to produce the final edited video.

\subsection{Multimodal Narrative Understanding}
% Short dramas, films, and other audiovisual productions are complex, multimodal systems centered around human characters. 
% To better understand video content, this paper proposes a novel framework inspired by filmmaking and screenwriting techniques. 
% By integrating audiovisual understanding models with large multimodal models, the framework converts video content into detailed textual representations, simulating human cognitive processing. 
Our framework combines audiovisual models with MLLMs to convert video content into detailed textual representations, simulating human cognition. 
It features dialogue analysis, character extraction, character-dialogue matching, video scene segmentation and memory module. 
% A memory module is introduced to handle long-form videos, maintaining intermediate representations and capturing inter-segment relationships.

\noindent\textbf{Dialogue Analysis}. 
Films and short dramas often feature dense dialogues, with key plot developments driven by character conversations. 
To capture this, we first apply ASR technology \cite{whisper}, though its transcripts are prone to errors from acoustic interference, background music, and character name ambiguities. 
To address these issues, we use optical character recognition (OCR) \cite{parseq} to extract subtitles directly from video frames, providing a reliable reference. Leveraging Doubao 1.5 pro \footnote{\url{https://seed.bytedance.com/en/special/doubao_1_5_pro}}, we align and correct ASR outputs based on OCR subtitles, significantly improving transcription accuracy. 
Additionally, we perform speaker diarization \cite{eend} to distinguish audio by speaker, enabling fine-grained narrative analysis.

\noindent\textbf{Character Extraction}. 
The goal of character extraction is to identify individuals in videos and build a database of their key information. While annotated data or metadata may sometimes be available, it is often missing and must be generated automatically.
To address this, we employ face detection \cite{arcface} and clustering: detecting faces in video frames, extracting facial embeddings, and clustering them into distinct identities.
Additionally, we enrich character profiles by extracting character's name or identity from drama dialogues. 
Integrating these information into the character database can facilitate deeper narrative comprehension.
% We finetune an MLLM that extracts character-related textual descriptions, such as names and identities, integrating these into the character database to facilitate deeper narrative comprehension.

\noindent\textbf{Character-Dialogue Matching}. 
To enable human-like video understanding, 
we introduce a specialized module that links identified characters to their dialogues and infers their relationships.
After applying speaker diarization, dialogue segments are initially unlabeled. 
% Simple heuristics, such as matching visible characters to active speakers, are used but often fail in complex cases like off-screen dialogues. 
% To enhance accuracy, we integrate multiple LLMs to jointly analyze visual, audio, and semantic cues. 
% Their outputs are fused via a scoring mechanism, with final results accepted only when confidence exceeds a high threshold.
To align the speaker with the spoken content, we input the character's facial image and dialogue information into Doubao 1.5 pro to reason and infer the speaker's identity. 
To ensure accuracy, we employ multiple inference iterations followed by a voting mechanism.

\noindent\textbf{Video Scene Segmentation}.
Video scene segmentation is crucial for dividing video content into coherent semantic units. 
Traditional methods often rely on low-level visual features, leading to fragmented outputs, especially in dramas with frequent shot transitions.
% To address this, we propose a three-stage approach: visual shot segmentation, local shot fusion, and global semantic refinement. 
% We first apply a visual segmentation model (e.g., TransNetV2 \citep{transnetv2}) to obtain shot-level boundaries. Neighboring shots are then clustered via a trained classifier to reduce fragmentation. 
% Finally, an MLLM refines segmentation globally, merging visually dissimilar but semantically related segments (e.g., close-ups and wide shots of the same scene).
To achieve this, we first apply AutoShot \cite{autoshot} to obtain shot-level boundaries and use Qwen2-VL \cite{wang2024qwen2} to merge shots with the same site (e.g. restaurant or office). 
Subsequently, we further utilize Gemini 2.0 flash to merge them belonging to the same event based on contextual information and offer.
Meanwhile, a description of the current event is generated.
This approach balances segmentation granularity and semantic completeness, ensuring transition fluency.

% \noindent\textbf{Comprehensive Caption}. 
% The Comprehensive Caption module generates holistic semantic summaries of films and dramas. 
% Unlike conventional methods that describe isolated segments, it integrates multimodal inputs from earlier analysis steps to produce coherent, context-aware narratives. 
% Specifically, key information such as characters, dialogues, relationships, and prior episode summaries are fed into an MLLM alongside the current video. 
% This ensures accurate, continuous descriptions that capture scene contexts, logical connections, and overall plot development.

\noindent\textbf{Memory}. 
The memory module supports processing requirements for long-form videos and multi-episode short dramas by organizing and managing data around two core dimensions: characters and narrative.
For characters, it maintains key information, including visual features, interpersonal relations, dialogues, and narrative trajectories across episodes or video segments. 
For the narrative dimension, it systematically records scene segmentation results and overall storyline progression. 
Additionally, this module interacts dynamically with other components, supports real-time updates, and enables multi-dimensional data retrieval, effectively accommodating storyline evolution and facilitating downstream analysis.

\begin{algorithm*}[t]
\caption{Highlight-based Automatic Video Editing Algorithm}
\begin{algorithmic}[1]
\State \textbf{Initialization:} Highlight rule $R_h$, Opening rule $R_o$, Ending rule $R_e$, Pruning rule $R_p$
\State \textbf{Input:} Video composed of $n$ scenes $V = (S_1, S_2, ..., S_n)$, LLM $L$, Top highlights number $k$
\State \textbf{Output:} Edited videos $V'$

\State $H = (h_1, h_2, ..., h_k) \gets L(R_h, V)$ \Comment{Extract highlight clips $H$ using LLM}
\State $H' \gets \text{sort}(H)$ \Comment{Sort highlight clips $H$ by scene score in descending order}
\State Initialize candidate editing boundary set $B \gets \{\}$

\For{each $h' \in H'[1:k]$}
\State $O = \{o_1, ..., o_m\} \gets L(R_o, h', V)$ \Comment{Select all suitable opening scenes for $h'$, $O\subseteq V$}
\State $E = \{e_1, ..., e_n\} \gets L(R_e, h', V)$ \Comment{Select all suitable ending scenes for $h'$, $E\subseteq V$}
\State $B \gets B \cup (O \times E)$ \Comment{Enumerate all openings-endings pairs by cartesian product}
\EndFor
% \State $B \gets \text{deduplicate}(B)$ \Comment{Remove duplicate boundary pairs}
\State Initialize edited video set $V' \gets \{\}$

\For{each $(o, e) \in B$}
\State $P = \{p_1, ..., p_l\} \gets L(R_p, V[o:e])$ \Comment{Get scenes to be pruned within $V[o, e]$, $P \subseteq V$}
\State $V'\gets V'\cup\text{Cut \& Splice}(V[o:e] \setminus P)$ \Comment{Cut and splice remaining scenes}
\EndFor
\State \Return $V'$
\end{algorithmic}
\label{alg:pseudo}
\end{algorithm*}

\subsection{Human-Inspired Editing Framework}
To explore better editing methods, we consult several professional advertisement editors and distill their strategies into our editing framework, which consists of three components: Highlight Detection, Opening \& Ending Selection, and Content Pruning. Algorithm \ref{alg:pseudo} provides a detailed illustration of the workflow of the editing algorithm.

\noindent\textbf{Highlight Detection}.
Highlight clips are crucial for producing engaging videos \cite{highlight2016, highlight2019, highlight2022}. 
In our framework, DeepSeek-R1 \cite{guo2025deepseek} is introduced to identify highlights.
Specifically, we abstract common highlight clips in short dramas and films into a set of rules and use R1 to evaluate whether each scene matches these rules (See Appendix \ref{app:highlight} for more details). 
Each scene is scored based on the number of matched patterns, while unmatched scenes are labeled as "general scenes" with zero scores. 
After scoring, adjacent non-zero-scored scenes are merged into "highlight clips", with the total score reflecting their importance.
% As illustrated in Figure \ref{fig:method}(c), highlight clip 1 consists of adjacent scenes 4, 5, and 6, whose individual scores sum up to 3 + 4 + 2 = 9.

\noindent\textbf{Opening \& Ending Selection}.
After highlight clips are detected, we select appropriate starting and ending scenes to form the video boundaries. 
For short drama advertisements, the opening aims to attract viewers' attention, while the ending leaves suspense or concludes climax to encourage clicks. 
Candidate starting scenes include preceding general scenes and the first scene of each highlight clip; ending candidates include following general scenes and the last scene of highlight clips. 
The GPT-4o evaluates these candidates (details in Appendix \ref{app:boundary}), and several opening-ending pairs are sampled to ensure variety.

\noindent\textbf{Content Pruning}.
Even after boundary selection, some redundant scenes may remain. 
To streamline the final video, the GPT-4o analyzes each general scene's relevance to the storyline. 
Less engaging scenes are pruned, ensuring a smooth narrative flow without unnecessary content. 
The result is an edited video with a captivating introduction, tightly packed highlights, and a suspenseful conclusion.

It is important to emphasize that we only present an editing framework. The framework itself is model-agnostic, allowing any required models to be replaced with more suitable alternatives depending on the practical context. Thus, our framework ensures both flexibility and adaptability across different applications. All prompts used in our framework are listed in Appendix \ref{app:templates}.

\section{DramaAD Dataset}
\subsection{Data Collection}
Firstly, we collect 30 popular Chinese short dramas from the internet.
Subsequently, we hire some advertisement editing experts to create advertisement videos for these short dramas as references. 
Additionally, we annotate the reference videos with the corresponding multimodal narration and editing timestamps to facilitate future research.
In this manner, we construct the proposed dataset, named DramaAD.

\begin{table*}[t]
% \vspace{-0.5em}
\centering
\resizebox{\textwidth}{!}{
\begin{tabular}{c|ccccc}
\toprule
% \multirow{2}{*}{\textbf{Method}} & \multicolumn{1}{c|}{ \textbf{Strategy} } & \multicolumn{3}{c|}{ \textbf{General Scenarios} } & \multicolumn{2}{c}{ \textbf{Advertising Scenarios} } \\
Dataset & Domain & Videos & Duration & Resolution & Reference Videos\\
% \cmidrule(lr){2-7}
% & Diversity & Smoothness & Engagement & VEI & Hook Rate & Suspense Rate \\ 
\midrule
YouCook2 \cite{youcook} & Cooking & 2K & 176h & - & - \\
InternVid \cite{internvid} & Open & 7.1M & 760.3Kh & 720P & -\\
Vript \cite{vript} & Open & - & 1.3Kh & 720P & - \\
% \midrule
AVE \cite{ave} & Movie & 5.5K & 207h & 720P-1080P & - \\
SkyScript-100M \cite{skyscript} & Drama & 6.6K & 2Kh & 720P & - \\
\textbf{DramaAD (Ours)} & \textbf{Drama} & 2.6K & 68h & 720P-1080P & \textbf{522} \\
\bottomrule
\end{tabular}}
\caption{Comparison between DramaAD and other related datasets.}
\label{tab:dataset}
\vspace{0.em}
\end{table*}

\subsection{Statistics and Comparison}
\begin{figure}[!t]
  \includegraphics[width=\linewidth]{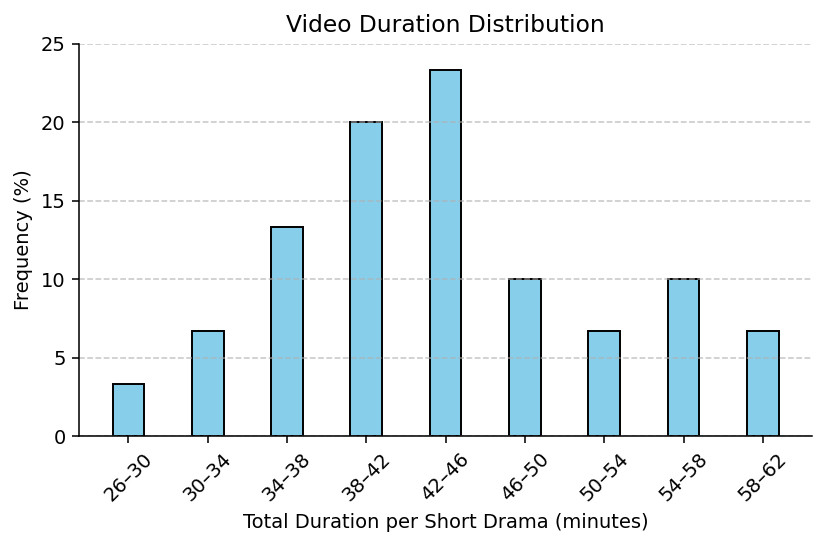}
  \vspace{0.em}
  \caption{The distribution of the total video duration for each short drama. The average duration is about 40 minutes.}
  \label{fig:duration}
  \vspace{0.em}
\end{figure}

DramaAD contains 3104 videos, of which 2582 are short drama videos and 522 are edited advertisement videos.
Compared to other datasets, our drama videos have 1080P resolution (the resolution of reference videos ranges from 720P to 1080P) and are in a 9:16 aspect ratio, which aligns better with the trend of increasing popularity of mobile short-form videos. The statistical information of the dataset and comparison with other datasets are presented in Figure \ref{fig:duration} and Table \ref{tab:dataset} seperately.
% Additionally, to the best of our knowledge, DramaAD is the first short drama dataset that includes reference videos for advertisement editing.

\subsection{Evaluation Metrics}
To conduct a comprehensive evaluation, we propose multiple quantitative metrics for assessing both the edited videos and the editing methods.
\noindent\textbf{Diversity.}
Diversity is utilized to measure the creativity of the editing methods.
Give raw footage, assuming the editing method generates $n$ edited videos $(V_1, V_2,...V_n)$, the diversity is calculated as follows:
\begin{align}
    \mathrm{Diversity}&=1-\frac{1}{\tbinom{n}{2}}\sum_{i<j}^{n}\mathrm{IoU}(V_i,V_j)
\end{align}

\noindent\textbf{Smoothness.}
Abrupt transitions are a common issue in automated video editing. 
We use a metric called smoothness to evaluate the overall coherence of the edited video. 
Smoothness is defined as the average uninterrupted duration, meaning the average number of minutes before an unnatural or jarring transition occurs.
\begin{equation}
    \mathrm{Smoothness}=\frac{\mathrm{Duration}(V)}{1+\mathrm{Interruption}(V)}
\end{equation}

\noindent\textbf{Engagement.}
Engagement captures the extent to which the edited video maintains viewers' attention.
Participants are instructed that they may either switch to 2× playback speed or skip forward using the progress bar if they find any part of the video uninteresting. 
We define Engagement as the proportion of the video watched at normal speed:
\begin{equation}
\mathrm{Engagement} = \frac{\mathrm{NormalPlayDuration}{}(V)}{\mathrm{Duration}(V)}
\end{equation}

\noindent\textbf{Viewing Experience Index.}
To provide a holistic view of video quality, we define the Viewing Experience Index (VEI) as the product of fluency and engagement:
\begin{equation}
    \mathrm{VEI}=\mathrm{Engagement}\times\mathrm{Smoothness}
\end{equation}

Besides general metrics, we propose two additional metrics specifically for advertising scenarios.

\noindent\textbf{Hook Rate.} 
It indicates whether viewers will pause to watch the video upon encountering it, reflecting the attractiveness of the video's opening.

\noindent\textbf{Suspense Rate.}
It measures whether viewers are motivated to continue watching due to the suspense created at the end of the video.

\section{Experiments}
\subsection{Experiment Settings}

\begin{table*}[t]
\vspace{-1.em}
\centering
\resizebox{\textwidth}{!}{
\begin{tabular}{c|c|ccc|cc}
\toprule
\multirow{2}{*}{\textbf{Method}} & \multicolumn{1}{c|}{ \textbf{Strategy} } & \multicolumn{3}{c|}{ \textbf{General Scenarios} } & \multicolumn{2}{c}{ \textbf{Advertising Scenarios} } \\
& Diversity & Smoothness & Engagement & VEI & Hook Rate & Suspense Rate \\ 
\midrule
Human (Golden) & 0.74 & 6.84 & 0.93 & 6.35 & 0.87 & 0.91 \\
\midrule
End2End (ASR) & \underline{0.75} & 0.65 & 0.82 & 0.54 & 0.64 & 0.47  \\
End2End (Narration) & 0.48 & 1.28 & \underline{0.88} & 1.14 & 0.62 & 0.52  \\
\midrule
\textbf{HIVE (ours)} & 0.66 & \underline{4.48} & \textbf{0.89} & \textbf{4.01} & \textbf{0.71} & \textbf{0.73} \\
a. w/o highlight & \textbf{0.78} & 4.42 & 0.62 & 2.74 & 0.65 & \underline{0.69} \\
b. w/o boundary & 0.54 & 4.17 & 0.81 & 3.38 & 0.28 & 0.30 \\
c. w/o pruning & 0.66 & \textbf{5.10} & 0.69 & \underline{3.51} & \underline{0.69} & 0.68 \\
\bottomrule
\end{tabular}}
\caption{Evaluation results on DramaAD dataset. The best result among the automatic editing methods is in \textbf{bold}, while the second best is \underline{underlined}.}
\label{tab:cut_results}
\vspace{-1.em}
\end{table*}

\noindent\textbf{Data.}
We sample 220 episodes from DramaAD dataset and select several complete storylines from these videos, which typically span approximately 10 episodes and last 20 to 30 minutes. 
These videos will serve as the raw footage for the editing process.

\noindent\textbf{Baselines and Reference.}
We compare our framework with the following baselines. 
\begin{itemize}
    \item[(1)] End2end editing method based on ASR-transcribed dialogue, which is a commonly used editing method. For fairness, we replace the prompts with the same requirements.
    \item[(2)] End2end editing method based on our video scene segmentation and narratives. Compared to methods in (1), we replace dialogues with scene narratives. 
For fairness, both methods are provided with the same editing expertise (See Appendix \ref{app:end2end} for more details).
    \item[(3)] Additionally, human-edited videos in the dataset will serve as the golden editing reference.
\end{itemize}

\noindent\textbf{Ablation Study Settings.}
We conduct ablation studies to analyze the contribution of the components in our framework.

\noindent\textbf{a. w/o highlight localization}
We directly view all scenes as starting and ending candidates.

\noindent\textbf{b. w/o boundary selection}
We randomly select starting and ending scenes from candidates.

\noindent\textbf{c. w/o pruning}
All clips from the opening to ending scenes are kept without filtering.

\subsection{Experimental Results}
\noindent\textbf{Main Results.}
As shown in Table \ref{tab:cut_results}, our proposed method outperforms end-to-end editing approaches in both general and advertisement scenarios.

ASR-based methods yield higher diversity, likely due to richer timestamp information in conversational data, but result in lower fluency, limiting their industrial applicability.
Comparing the two end-to-end methods, our video understanding framework improves fluency through semantic-based shot aggregation, albeit at the expense of some diversity.

\noindent\textbf{Ablation Analysis.}
Ablation studies reveal that removing highlight detection increases diversity but lowers the engagement ratio. The increase in diversity stems from the model’s ability to freely select editing ranges across all scenes. 
Randomly selecting opening and ending scenes significantly weakens advertisement effectiveness and slightly decreases engagement.
Omitting scene pruning leads to a lower proportion of engaging content in the final video, but it reduces abrupt visual transitions, thereby improving smoothness. \footnote{However, since no additional processing is applied to the ending effects when stitching multiple episodes, abrupt transitions may still occur occasionally.}

% Ablation studies reveal that removing highlight detection increases diversity but leads to a decrease in engagement. 
% The increase in diversity stems from the model’s ability to freely select editing ranges across all scenes. 
% However, without explicit highlight detection, the resulting clips are less likely to contain sufficiently engaging moments, ultimately compromising viewer engagement.

% If the opening and ending scenes are not filtered, the edited video may begin or end at arbitrary points, making it less effective at capturing the viewer’s attention and sustaining engagement. 
% This can significantly reduce user retention and lead to a sharp decline in key advertising metrics.

% Omitting scene pruning leads to a lower proportion of engaging content in the final video, but it helps reduce abrupt visual transitions, thereby improving smoothness. 
% \footnote{However, since no additional processing is applied to the ending effects when stitching multiple episodes, abrupt transitions may still occur occasionally.}

\section{Conclusion}
In this work, we present a novel human-inspired framework for automatic video editing, addressing key limitations of existing ASR-based approaches. 
By integrating multimodal narrative understanding and decomposing the editing process into intuitive sub-tasks, our method achieves superior editing quality with improved interpretability. 
To further advance research in this area, we introduce DramaAD, the first large-scale short drama dataset with reference advertisement edits, providing valuable resources for both academia and industry. 
Experimental results demonstrate that our approach significantly narrows the gap between automated and human editing.
We hope our work will advance research in the field of automatic video editing.

\section*{Limitations}
\begin{itemize}
    \item[1.] The proposed method is currently not compatible with editing techniques such as flashbacks or non-linear storytelling.
    \item[2.] Since video scene segmentation divides the original video into coarser units compared to shot segmentation, it may reduce the flexibility of editing.
    \item[3.] Compared to human editing, the proposed method still falls short in terms of metrics such as diversity and smoothness. We hope this work can contribute to advancing research in this area.
\end{itemize}

\section*{Ethics Statement}
We strictly follow the ACL Code of Ethics and comply with all guidelines during the research.
Our work presents two potential ethical concerns: data source and human editing services.

\noindent\textbf{Data Source.}
All short drama data in our dataset are authorized by the copyright holders, allowing us to conduct academic research based on these videos.

\noindent\textbf{Human Editing Services.}
In constructing the dataset, we enlisted experienced editing professionals to generate the reference edited videos. Each video took around 30 minutes to complete, and the experts were compensated fairly based on local rates.

\section*{Acknowledgments}
This work was supported in part by the grants from National Natural Science Foundation of China (No.62222213, U22B2059).

\bibliography{custom}

\newpage
\appendix
\section{Prompt Templates}
\label{app:templates}
Here we present our detailed instruction templates which have been translated into English.

\subsection{Dialogue Analysis}
We instruct LLM to correct ASR-transcribed dialogue with OCR, the prompt template is shown in Table \ref{tab:correct_asr}.

\subsection{Comprehensive Caption}
Table \ref{tab:caption} shows the details of the scene description we obtained using MLLM as well as multidimensional information.

\subsection{Highlight Detection}
\label{app:highlight}
We defined the highlight scenes of the short dramas according to the gender of their target audience.
Table \ref{tab:hl_male} presents the highlight rules for the male audience, while table \ref{tab:hl_female} presents those for the female audience.
Table \ref{tab:hl} shows the instructions for highlight detection.

\subsection{Opening \& Ending Selection}
\label{app:boundary}
Table \ref{tab:boundary} is about selecting an appropriate opening and ending.
Here our prompt is to serve advertising purposes.

\subsection{Content Pruning}
\label{app:prune}
Please see Table \ref{tab:prune} for more details.

\subsection{End2End Editing}
\label{app:end2end}
End2End Editing instruction is shown in Table \ref{tab:end2end_editing}, and the ASR version is presented in Table \ref{tab:end2end_asr_editing}.

\begin{table*}
\begin{tcolorbox}[colback=gray!10,% gray background
                  colframe=black,% black frame color
                  width=1\textwidth,% Use 100% of text width
                  arc=1 mm, %auto outer arc,
                  boxrule=0.5pt,
                  title=Instruction for Correct ASR with OCR,
                  fontupper=\small
                 ]
\textbf{\#\# Role}\\
You are an expert in video dialogue content recognition. You're provided with the ASR (Automatic Speech Recognition) results and OCR (Optical Character Recognition) results from the same video, both with associated timestamp information. Your task is to cross-reference these two inputs to correct the ASR results to produce the most accurate speaker identification and dialogue content. The time information in the ASR must remain unchanged. \\
\textbf{\#\# Input}\\
- ASR: Includes timestamps, speaker identification, and spoken content. Due to potential inaccuracies in detection, both the speaker and the dialogue content might have errors and require correction.\\
- OCR: Includes timestamps and recognized text from the video's visuals. While OCR text is generally accurate, it may occasionally contain irrelevant visual noise (e.g., text from objects or advertisements), which requires careful filtering.\\
\textbf{\#\# Output}\\
- Corrected ASR:  Includes timestamps, corrected speaker identification, and dialogue content. Time remains unchanged from the original ASR. Corrections are made using OCR where inaccuracies in speaker or content are evident, while irrelevant OCR text is excluded. Speaker labels are adjusted only for clear mistakes to ensure accuracy.\\
\textbf{\#\# Requirements}\\
1. Cross-reference the timestamp in OCR with the timestamp in ASR to adjust the ASR speaker and spoken dialogue content.\\
2. Correct typical errors such as phonetic spelling issues in ASR by using OCR text with matching timestamps.\\
3. Ensure that the corrected ASR speaker and content align logically within the context of the video.\\
4. Maintain the original timestamp information from the ASR, regardless of corrections made.\\
5. Filter out irrelevant OCR content such as noise, and avoid overwriting ASR with unrelated visual information.\\
% \textbf{\#\# Examples}\\
% ASR: \\
% 00:01:05 Speaker A: She's like a store on the rest.\\
% 00:01:10,Speaker B: Yeah, it's a best to fine that.\\
% OCR:\\
% 00:01:05 She’s like a star on the rise.\\
% 00:01:10 Yeah, it’s best to find that. \\
% 00:01:15 Advertisement: Great offers waiting for you!\\
% Corrected ASR:\\
% 00:01:05 Speaker A: She’s like a star on the rise.\\
% 00:01:10"Speaker B: Yeah, it’s best to find that.\\
ASR:\\
\textbf{\{ASR\}}\\
OCR:\\
\textbf{\{OCR\}}\\

\end{tcolorbox}
\caption{Instruction template for Correct ASR with OCR}
\label{tab:correct_asr}
\end{table*}

\begin{table*}
\begin{tcolorbox}[colback=gray!10,% gray background
                  colframe=black,% black frame color
                  width=1\textwidth,% Use 100% of text width
                  arc=1 mm, %auto outer arc,
                  boxrule=0.5pt,
                  title=Instruction for Comprehensive Caption,
                  fontupper=\small
                 ]
\textbf{\#\# Role}\\
You are a professional film and drama captioning expert. Your task is to produce accurate and coherent narrative descriptions of key scenes in a video segment by integrating multimodal inputs while maintaining logical plot continuity.
 \\
\textbf{\#\# Input}\\
- Current Video Segment:
The specific video clip or sampled frames corresponding to the segment that needs to be summarized.\\
- Character Information: A list of characters appearing in this segment and their roles in the drama. \\
- Dialogue: A transcript of all dialogues spoken within this scene, including timestamps and speaker attributions. \\
- Previous Segment Context:
A concise overview of important events or character dynamics from the previous segment.\\
\textbf{\#\# Output}\\
- Comprehensive Description:\\
A coherent and context-aware summary of the segment that integrates characters, dialogue, and contextual information to reflect its significance within the broader narrative. Key events, character interactions, and logical plot connections should be naturally framed, avoiding verbatim dialogue and focusing on emotional and story progression.\\
\textbf{\#\# Requirements}\\
1. Holistic Scene Summarization:Generate a coherent narrative that captures key events, character motivations, relationships, and emotional flow while emphasizing the scene's role in advancing the plot or developing character arcs. \\
2. Continuity Maintenance: Ensure the summary remains consistent with the Previous Episode Summary and Current Episode Summary, connecting past interactions or unresolved conflicts to the current scene within the broader storyline.\\
3. Incorporate Dialogue and Key Interactions: Integrate key dialogues naturally into the description to highlight important developments, reframing them fluently into the narrative without listing them verbatim. \\
4. Context Sensitivity: Use all provided inputs (Character Information, Dialogue, Previous Segment Context and Current Video Segment) to create a summary aligned with both scene details and the overall plot progression. \\
Current Video Segment:\\
\textbf{\{Current Video Segment\}}\\
Character Information:\\
\textbf{\{Character Information\}}\\
Dialogue:\\
\textbf{\{Dialogue\}}\\
Previous Segment Context:\\
\textbf{\{Previous Segment Context\}}\\
\end{tcolorbox}
\caption{Instruction template for Comprehensive Caption}
\label{tab:caption}
\end{table*}

\begin{table*}
\begin{tcolorbox}[colback=gray!10,% gray background
                  colframe=black,% black frame color
                  width=1\textwidth,% Use 100% of text width
                  arc=1 mm, %auto outer arc,
                  boxrule=0.5pt,
                  title=Highlight Definition and Scoring Rules for Male Audience,
                  fontupper=\small
                 ]
\textbf{Category 1 – Modern Male-Lead Stories}\\
This category usually focuses on a male protagonist who hides his true identity or abilities, playing the underdog while secretly being powerful. \\
\textbf{Common plots include}:\\
- Unknown Hero Rises: A seemingly ordinary man suddenly enters the scene, shaking up the world around him.\\
- Return and Revenge: The male lead is forced to stay away for some reason. When he returns, he finds his wife and children have suffered, prompting him to take revenge.\\
\textbf{Typical Highlight Moments}:\\
- First Encounter with Female Lead: Their first meeting usually involves intense conflict and serves as an early mini-climax. (3 points)\\
- Mocked or Doubted: A classic setup where the hero is looked down upon, making the audience eager to see how he proves everyone wrong. (3 points)\\
- Praised by Powerful Figures: When villains attack the hero, someone with a strong background—like a senior or powerful ally—steps in to support him, subtly hinting at his true status. (3 points)\\
- Hero Saves the Beauty: A timeless scene—rescuing the female lead when she's in trouble, avenging his wife, or helping her through a career crisis. (3 points) \\
- Beautiful Woman Appears Out of Nowhere: Whatever the hero wants, it conveniently shows up. For example, an old man whose life the hero saved offers his daughter in marriage, or after the ex-wife scorns him, new admirers suddenly appear. (3 points) \\
\textbf{If none of the above apply, you can choose}:\\
- Betting Scenes: Moments involving wagers or dares. (2 points)\\
- Running into an Ex: Meeting an ex-girlfriend, wife, etc. (2 points)\\
- Large Crowds: Scenes like auctions, competitions, or any event involving many extras. (2 points)\\
\textbf{Category 2 – Historical Male-Lead Stories}\\
These stories often feature a male protagonist who uses memories from a past life or modern-day knowledge to continuously rise in status. They can be split into time-travel or reincarnation subgenres.\\
\textbf{Typical Highlight Moments}:\\
- First Pot of Gold: Usually the first big moment in the series—such as using modern knowledge to earn money. (3 points)\\
- Mocked, Then Fights Back: Again, a classic "looked down upon but rises up" setup. (3 points)\\
- First Meeting with Female Lead: This encounter usually comes with dramatic tension. (3 points)\\
\textbf{Category 3 – General Highlights}\\
\textbf{These highlight moments apply broadly across all story types}:\\
- Major Climax or Turning Point: This could involve a key character's death, a huge secret revealed, or any major plot shift. (2 points)\\
- Emotional Resonance: Moments that trigger strong emotions—whether sadness, joy, shock, or fear—that stay with the audience. (1 point)\\
- Cultural Impact: Scenes that spark social discussion or shift public attitudes. (1 point)\\
- Cliffhanger: Endings that leave viewers eagerly awaiting the next episode. (2 points)\\
- Key Plot Twist: Includes major developments like the heroine’s identity being revealed or the story’s main conflict emerging. (2 points)
\end{tcolorbox}
\caption{Highlight scene definition and scoring rules for the male audience.}
\label{tab:hl_male}
\end{table*}

\begin{table*}
\begin{tcolorbox}[colback=gray!10,% gray background
                  colframe=black,% black frame color
                  width=1\textwidth,% Use 100% of text width
                  arc=1 mm, %auto outer arc,
                  boxrule=0.5pt,
                  title=Highlight Definition and Scoring Rules for Female Audience,
                  fontupper=\small
                 ]
\textbf{Category 1 – Modern Female-Lead Stories}
These stories are told mainly from the female protagonist's perspective. They typically fall into two types:\\
- Sweet Romance: Often follows a "married first, fall in love later" plot, with plenty of sweet, heartwarming moments sprinkled throughout.\\
- Revenge \& Glow-Up: The heroine is divorced by her ex-husband, only to thrive and shine even brighter afterward.\\
\textbf{Common Highlight Moments}:\\
- Unexpected Accidents: For example, the heroine gets into a car accident. (3 points)\\
- Arguments \& Face-Offs: Such as the heroine confronting a rival female character. (3 points)\\
- Divorce Scenes: Like the heroine asking the male lead for a divorce. (3 points)\\
- Identity Reveals or Secrets Uncovered: For instance, it turns out the heroine is a highly skilled expert. (2 points)\\
- Humor \& Comedy: Scenes with funny twists, e.g., the heroine making the CEO male lead ride an electric scooter. (2 points)\\
- Violence \& Intensity: High-drama moments like slapping scenes. (3 points)\\
- Flirty \& Ambiguous Moments: Light teasing or suggestive scenes that keep viewers hooked. (3 points)\\
- First Encounter Between Leads: Usually the first mini-climax of the show. (3 points)\\
- One-Night Stand/Pregnancy Plotlines: A one-night stand leads to future encounters filled with potential drama. (3 points)\\
- Heroine Actively Pursues the Male Lead: "Chasing love" moments are often the sweetest parts. (3 points)\\
- The heroine in Trouble, Rescued by Male Lead: Typically involves the heroine being mocked, drugged, attacked, or assassinated—only for the male lead to appear heroically. (3 points)\\
- Heroine's Suffering Moments: Classic melodrama setup, where she is initially oppressed before rising up. (3 points)\\
- Heroine’s Counterattack: She fights back against vicious rivals, ex-husbands, or antagonists. (3 points)\\
\textbf{If none of the above apply, you can choose}:\\
- Sweet Interactions Between Leads: Romantic or affectionate scenes. (2 points)\\
- Male Lead Protecting or Supporting the Heroine: Either defending her or helping with her career. (2 points)\\
- Pregnancy Reveal: When either lead learns about the pregnancy. (2 points)\\
- Heroine Pressured to Divorce: Scenes where the ex-husband or in-laws force her to divorce. (2 points)\\
\textbf{Category 2 – Historical Female-Lead Stories}\\
These stories usually feature a heroine who uses memories from a past life or modern knowledge to outsmart rivals and win true love.\\
\textbf{Common Highlight Moments:}\\
- Pre/Post Time-Travel or Rebirth Scenes: The opening explains the setup, drawing viewers in. (3 points)\\
- Heroine’s Counterattack: She strikes back at jealous female rivals, villains, or underlings. (3 points)\\
- Heroine in Peril: Trapped or framed by villains, followed by a rescue. (3 points)\\
\textbf{If none of the above apply, you can choose}:\\
- Sweet Interactions Between Leads: Hugging, cheek touching, or other intimate moments. (2 points)\\
- Villain’s First Appearance: Introduction of key antagonists. (2 points)\\
\textbf{Category 3 – General Highlights}\\
\textbf{These highlight moments apply broadly across all story types}:\\
- Major Climax or Turning Point: This could involve a key character's death, a huge secret revealed, or any major plot shift. (2 points)\\
- Emotional Resonance: Moments that trigger strong emotions—whether sadness, joy, shock, or fear—that stay with the audience. (1 point)\\
- Cultural Impact: Scenes that spark social discussion or shift public attitudes. (1 point)\\
- Cliffhanger: Endings that leave viewers eagerly awaiting the next episode. (2 points)\\
- Key Plot Twist: Includes major developments like the heroine’s identity being revealed or the story’s main conflict emerging. (2 points)
\end{tcolorbox}
\caption{Highlight scene definition for the female audience.}
\label{tab:hl_female}
\end{table*}

\begin{table*}
\begin{tcolorbox}[colback=gray!10,% gray background
                  colframe=black,% black frame color
                  width=1\textwidth,% Use 100% of text width
                  arc=1 mm, %auto outer arc,
                  boxrule=0.5pt,
                  title=Instruction for Highlight Detection,
                  fontupper=\small
                 ]
\textbf{\#\# Role}\\
You are a professional short drama editor with a deep understanding of plot development and audience engagement. Your task is to evaluate the highlight level of each scene based on specific scoring guidelines and return the results accordingly.\\
\textbf{\#\# Input}\\
- Drama Title: The name of the short drama.\\
- Target Audience Gender: The primary gender demographic of the audience.\\
- Scene Fragments: Several scene descriptions from a specific episode.\\
\textbf{\#\# Output} \\
You should output the score for each scene in sequence, formatted as a JSON list, wrapped with <result> tags. The format should look like this:\\
<result> \\
\text{[}\\
  \text{\{"episode": 1, "scene\_id": 1, "reason": "Your justification for the score of this scene.", "score": 3\},}\\
  \text{\{"episode": 1, "scene\_id": 2, "reason": "Your justification for the score of this scene.", "score": 0\},}\\
  ...\\
\text{]}\\
</result>\\
Explanation:\\
- episode: The episode number.\\
- scene\_id: The scene number within that episode.\\
- score: The score assigned to the scene.\\
- reason: The rationale behind the score.\\
Please note: The input scene fragments may not start from Episode 1. Make sure to keep the episode and scene numbers consistent in your output.\\
\textbf{\#\# Requirements}\\
1. I will provide you with editing insights that summarize key elements of popular and engaging plotlines. Based on these guidelines, please evaluate each scene as follows:\\
2. If a scene matches one or more of the provided criteria, add up the corresponding points.\\
3. If the scene does not meet any criteria or feels plain, assign a score of 0.\\

\textbf{\{Highlight definition and scoring rules (dependent on audience gender)\}}\\

\textbf{\{Scene narration of episode 1 $\sim$ T\}}\\
\end{tcolorbox}
\caption{Instruction template for highlight detection.}
\label{tab:hl}
\end{table*}
\begin{table*}
\begin{tcolorbox}[colback=gray!10,% gray background
                  colframe=black,% black frame color
                  width=1\textwidth,% Use 100% of text width
                  arc=1 mm, %auto outer arc,
                  boxrule=0.5pt,
                  title=Instruction for Opening \& Ending Selection,
                  fontupper=\small
                 ]
\textbf{\#\# Role}\\
You are a professional short drama editor with a deep understanding of narrative structure and audience engagement. Your goal is to create captivating edited videos by highlighting the most exciting scenes. For simplicity, the editing strategy is as follows:\\
- First, select the pre-identified highlight scenes as the core selling points.\\
- Then, choose suitable starting and ending scenes to wrap around these highlights.\\
In this task, the highlight scenes are already marked. You need to analyze the drama and select the best start and end scenes accordingly.\\
\textbf{\#\# Input}\\
- Drama Title: The name of the short drama.\\
- Target Audience Gender: The primary gender demographic of the audience.\\
- Drama Plot: The storyline of the drama, is composed of multiple scenes. Each scene contains an episode ID, a scene ID, and a description of its visual content. Some scenes are labeled as follows:\\
- <Highlight>: Key exciting scenes that serve as the main attraction in the edited video.\\
- <Optional Start>: Scenes pre-selected as potential starting points.\\
- <Optional End>: Scenes pre-selected as potential ending points.\\
Some scenes may carry multiple tags at the same time.\\
\textbf{\#\# Output} \\
For each scene labeled as <Optional Start> or <Optional End>, you need to decide whether it is suitable to be used as the start or end of the video.\\\
Output your decision as a JSON list wrapped in <result> tags, like this:\\
<result> \\
\text{[}\\
\{"episode": 1, "scene\_id": 1, "thought": "Your reason about if this scene is a suitable start or end scene.", "starting": true, "ending": false\},\\
\{"episode": 1, "scene\_id": 2, "thought": "Your reason about if this scene is a suitable start or end scene.", "starting": false, "ending": false\},\\
\{"episode": 2, "scene\_id": 4, "thought": "Your reason about if this scene is a suitable start or end scene.", "starting": false, "ending": true\},\\
  ...\\
\text{]}\\
</result>\\
Explanation:\\
- episode: episode number.\\
- scene\_id: scene number within that episode.\\
- thought: your reasoning and decision for each scene.\\
- starting: true if suitable as the start scene; otherwise false.\\
- ending: true if suitable as the end scene; otherwise false.\\
\textbf{\#\# Requirements}\\
\textbf{For Selecting Start Scenes:}\\
You should analyze the plot and consider:\\
- Audience Engagement: The opening scene should immediately grab attention and draw viewers in.\\
- Clarity: Avoid starting with scenes that rely heavily on prior plot context, such as ones mid-event, which might confuse new viewers.\\
- Introduction Scenes: Scenes that introduce characters, setting, or plot premise can be good starting points.\\
\textbf{For Selecting End Scenes}:\\
You should analyze the plot and consider:\\
- Relevance: Avoid ending on scenes unrelated to the highlight, as they may transition to a new, less exciting storyline.\\
- Neutral Endings: If a scene doesn't strongly affect the viewing experience, either way, it can still be chosen as an ending.\\
- Suspense: Prefer ending on scenes that leave a sense of suspense, encouraging viewers to click to find out what happens next.\\
- Complete Story Arc: It is also acceptable to end where a plot arc is fully wrapped up, giving a satisfying sense of closure while showcasing the highlight.\\

\textbf{\{Scene narration of episode 1 $\sim$ T with tags\}}\\
\end{tcolorbox}
\caption{Instruction template for Opening \& Ending Selection.}
\label{tab:boundary}
\end{table*}
\begin{table*}
\begin{tcolorbox}[colback=gray!10,% gray background
                  colframe=black,% black frame color
                  width=1\textwidth,% Use 100% of text width
                  arc=1 mm, %auto outer arc,
                  boxrule=0.5pt,
                  title=Instruction for Content Pruning,
                  fontupper=\small
                 ]
\textbf{\#\# Role}\\
You are a professional short drama editor with a deep understanding of storyline structure. Your goal is to craft engaging and cohesive edited videos. For this task, a relatively attractive video segment has already been pre-selected for you. Your job is to analyze the storyline and remove redundant scenes to make the plot more concise and gripping.\\
\textbf{\#\# Input}\\
- Drama Title: The name of the short drama.\\
- Target Audience Gender: The primary gender demographic of the audience.\\
- Drama Plot: The storyline is composed of multiple scenes. Each scene includes an episode ID, a scene ID, and a description of the visual content.\\
Additionally, each scene is labeled as either <Highlight Scene> or <General Scene>:\\
- <Highlight Scene> refers to key moments designed to attract viewers and evoke strong emotions. These are the main selling points of the edited video.\\
- <General Scene> refers to relatively plain or transitional scenes, which are candidates for possible removal.\\
\textbf{\#\# Output} \\
For every <General Scene>, you need to decide whether it can be deleted. Generate your decision as a JSON list wrapped in <result> tags, like this:\\
<result> \\
\text{[}\\
\{"episode": 1, "scene\_id": 1, "thought": "Your reasoning on whether this scene should be kept or removed.", "delete": false\},\\
\{"episode": 1, "scene\_id": 3, "thought": "Your reasoning on whether this scene should be kept or removed.", "delete": true\},\\
\{"episode": 2, "scene\_id": 2, "thought": "Your reasoning on whether this scene should be kept or removed.", "delete": false\},\\
  ...\\
\text{]}\\
</result>\\
Explanation:\\
- episode: Episode number.\\
- scene\_id: Scene number within that episode.\\
- thought: Your reasoning about whether this scene can be removed.\\
- delete: true if the scene should be removed; false if it should be kept.\\
\textbf{\#\# Requirements}\\
1. Only remove scenes labeled as <Regular Scene>. Scenes marked as <Highlight Scene> must never be deleted.\\
2. Do not remove the first or last scene, even if labeled as <Regular Scene>.\\
3. If there are no deletable scenes, simply output an empty list inside the <result> tag.\\
4. Maintain the original episode and scene\_id numbers as provided.\\
5. Be cautious when removing scenes. Unnecessary deletions may affect the coherence of the storyline, negatively impacting the viewing experience.\\

\textbf{\{Scene narration of episode 1 $\sim$ T with tags\}}\\
\end{tcolorbox}
\caption{Instruction template for Content Pruning.}
\label{tab:prune}
\end{table*}
\begin{table*}
\begin{tcolorbox}[colback=gray!10,% gray background
                  colframe=black,% black frame color
                  width=1\textwidth,% Use 100% of text width
                  arc=1 mm, %auto outer arc,
                  boxrule=0.5pt,
                  title=Instruction for End2End Editing,
                  fontupper=\small
                 ]
\textbf{\#\# Role}\\
You are a professional short drama editor with a deep understanding of plot development and audience engagement. \\
\textbf{\#\# Input}\\
- Drama Title: The name of the short drama.\\
- Target Audience Gender: The primary gender demographic of the audience.\\
- Scene Fragments: Several scene descriptions from a specific episode.\\
\textbf{\#\# Requirements}\\
1. You need to fully understand the content of the input short drama, analyzing the plot and characters to grasp the overall storyline.\\
2. Based on the scene descriptions, you should edit the drama by preserving the key plot points while ensuring the final cut remains coherent and smooth.\\
3. Pay special attention to retaining highlight moments within the scenes, as these can significantly enhance the appeal of the edited video. However, always prioritize the overall narrative flow when selecting which highlight moments to include.\\
I will provide highlight definitions and scoring rules for you.\\

\textbf{\{Highlight definition and scoring rules (dependent on audience gender)\}}\\

Besides highlights, good opening and ending scenes are also crucial.\\
\textbf{For Selecting Start Scenes:}\\
You should analyze the plot and consider:\\
- Audience Engagement: The opening scene should immediately grab attention and draw viewers in.\\
- Clarity: Avoid starting with scenes that rely heavily on prior plot context, such as ones mid-event, which might confuse new viewers.\\
- Introduction Scenes: Scenes that introduce characters, setting, or plot premise can be good starting points.\\
\textbf{For Selecting End Scenes}:\\
You should analyze the plot and consider:\\
- Relevance: Avoid ending on scenes unrelated to the highlight, as they may transition to a new, less exciting storyline.\\
- Neutral Endings: If a scene doesn't strongly affect the viewing experience, either way, it can still be chosen as an ending.\\
- Suspense: Prefer ending on scenes that leave a sense of suspense, encouraging viewers to click to find out what happens next.\\
- Complete Story Arc: It is also acceptable to end where a plot arc is fully wrapped up, giving a satisfying sense of closure while showcasing the highlight.\\
\textbf{\#\# Output} \\
When generating the output, you should reflect on these requirements and devise an appropriate editing strategy. After careful consideration, select the scenes you believe are suitable for the final cut and include them in a list. Remember to use <result> as the delimiter and present the edited script in the following format:\\
<result> \\
\text{[}\\
\{"episode":1, "scene\_id": 0, "thought": Your justification for choosing this scene. \}\\
\{"episode":1, "scene\_id": 2, "thought": Your justification for choosing this scene. \}\\
  ...\\
\text{]}\\
</result>\\
Explanation:\\
- episode: episode number.\\
- scene\_id: scene number within that episode.\\
- thought: your reasoning and decision for each scene.\\

\textbf{\{Scene narration of episode 1 $\sim$ T\}}\\
\end{tcolorbox}
\caption{Instruction template for End2End Editing.}
\label{tab:end2end_editing}
\end{table*}

\begin{table*}
\begin{tcolorbox}[colback=gray!10,% gray background
                  colframe=black,% black frame color
                  width=1\textwidth,% Use 100% of text width
                  arc=1 mm, %auto outer arc,
                  boxrule=0.5pt,
                  title=Instruction for End2End Editing with ASR,
                  fontupper=\small
                 ]
\textbf{\#\# Role}\\
You are a professional short drama editor with a deep understanding of plot development and audience engagement. \\
\textbf{\#\# Input}\\
- Drama Title: The name of the short drama.\\
- Target Audience Gender: The primary gender demographic of the audience.\\
- Dialogue Information: Dialogue information includes the start time, end time, and the content of each utterance.\\
\textbf{\#\# Requirements}\\
1. You need to fully understand the content of the input short drama, analyzing the plot and characters to grasp the overall storyline.\\
2. Based on the scene descriptions, you should edit the drama by preserving the key plot points while ensuring the final cut remains coherent and smooth.\\
3. Pay special attention to retaining highlight moments within the scenes, as these can significantly enhance the appeal of the edited video. However, always prioritize the overall narrative flow when selecting which highlight moments to include.\\
I will provide highlight definitions and scoring rules for you.\\

\textbf{\{Highlight definition and scoring rules (dependent on audience gender)\}}\\

Besides highlights, good opening and ending scenes are also crucial.\\
\textbf{For Selecting Opening:}\\
You should analyze the plot and consider:\\
- Audience Engagement: The opening shots should immediately grab attention and draw viewers in.\\
- Clarity: Avoid clips that rely heavily on prior plot context, such as ones mid-event, which might confuse new viewers.\\
- Introduction Shots: Shots that introduce characters, setting, or plot premise can be good starting points.\\
\textbf{For Selecting Ending}:\\
You should analyze the plot and consider:\\
- Relevance: Avoid endings that are unrelated to the highlight, as they may transition to a new, less exciting storyline.\\
- Neutral Endings: If a shot doesn't strongly affect the viewing experience, either way, it can still be chosen as an ending.\\
- Suspense: Prefer endings that leave a sense of suspense, encouraging viewers to click to find out what happens next.\\
- Complete Story Arc: It is also acceptable to end where a plot arc is fully wrapped up, giving a satisfying sense of closure while showcasing the highlight.\\
\textbf{\#\# Output} \\
In the output phase, you can consider these requirements and devise an editing strategy first. Finally, save the segments you deem suitable for editing into a list, and output your decision as a JSON list wrapped in <result> tags, like this:\\
<result> \\
\text{[}\\
\{"episode":1, "start\_time": 0, "end\_time": 14, "thought": Your justification for choosing this clip. \}\\
\{"episode":1, "start\_time": 18, "end\_time": 125, "thought": Your justification for choosing this clip. \}\\
  ...\\
\text{]}\\
</result>\\
Explanation:\\
- episode: episode number.\\
- start\_time: starting time of selected clip.\\
- end\_time: ending time of selected clip.\\
- thought: your justification for choosing this clip.\\

\textbf{\{ASR information of episode 1 $\sim$ T\}}\\
\end{tcolorbox}
\caption{Instruction template for End2End Editing with ASR.}
\label{tab:end2end_asr_editing}
\end{table*}

\end{document}